\documentclass[10pt,twocolumn,letterpaper]{article}

\usepackage{wacv}
\usepackage{times}
\usepackage{epsfig}
\usepackage{graphicx}
\usepackage{amsmath}
\usepackage{amssymb}
\usepackage{url}
\usepackage{booktabs}
\usepackage{bbold}
\usepackage[lined,ruled,linesnumbered]{algorithm2e}
\usepackage{xcolor}
\usepackage{verbatim}
\usepackage{enumitem}

\wacvfinalcopy 

\def\wacvPaperID{****} 

\ifwacvfinal\fi
\begin{document}

\title{Reducing Geographic Performance Differentials for Face Recognition}

\author{Martins Bruveris \\
Onfido, UK \\
{\tt\small martins.bruveris@onfido.com}
\and
Pouria Mortazavian \\
Onfido, UK \\
{\tt\small pouria.mortazavian@onfido.com}
\and
Jochem Gietema \\
Onfido, UK \\
{\tt\small jochem.gietema@onfido.com}
\and
Mohan Mahadevan \\
Onfido, UK \\
{\tt\small mohan.mahadevan@onfido.com}
}

\maketitle
\ifwacvfinal\thispagestyle{empty}\fi

\begin{abstract}
As face recognition algorithms become more accurate and get deployed more widely, it becomes increasingly important to ensure that the algorithms work equally well for everyone. We study the geographic performance differentials---differences in false acceptance and false rejection rates across different countries---when comparing selfies against photos from ID documents. 
We show how to mitigate geographic performance differentials using sampling strategies despite large imbalances in the dataset. Using vanilla domain adaptation strategies to fine-tune a face recognition CNN on domain-specific doc-selfie data improves the performance of the model on such data, but, in the presence of imbalanced training data, also significantly increases the demographic bias. We then show how to mitigate this effect by employing sampling strategies to balance the training procedure.
\end{abstract}

\section{Introduction}






Face recognition algorithms have enjoyed unprecedented performance improvements over the past few years. This has lead to multiple new commercial applications of facial recognition software, particularly in remote settings, from legal identity verification using a government issued ID to device authentication when accessing one's phone. 

Recent studies have raised concerns about commercial face recognition being ``biased'' with respect to certain demographic groups, demonstrating significantly different accuracy when applied to different demographic groups \cite{racialfw2019}. Such bias can result in mistreatment of certain demographic groups, by either exposing them to a higher risk of fraud, or by making access to services more difficult. Hence it is of utmost importance for any real world application to address such performance differentials among different demographic groups. It is commonly accepted that one of the major sources of performance differential in modern face recognition engines based on deep convolutional neural networks is demographic imbalances in the training data used to train these engines. 

We study the effect of demographic imbalance in training data in the setting of remote identity recognition where a selfie photo of the user's face is compared to a picture of an identity document. This is a cross-domain, one-to-one comparison scenario, in which the two images are from two different domains. We use a large-scale in-house dataset of image pairs of selfies and document images with fine-grained country and gender labels to quantify the performance differential in terms of false acceptance and false rejection rates within and across different geographic groups. We explore training strategies to minimise the performance differential among different demographic groups in presence of extreme data imbalance, where the smallest demographic group constitutes only 0.5\% of the total training data.



The contributions of this paper are as follows: 
\begin{itemize}[nosep]
    \item We present the first large-scale study of the effects of domain transfer by fine-tuning on selfie-doc data, where  images of ID documents are used as opposed to digital images extracted from the NFC chip.
    \item We present the first study of selfie-doc face matching across a large range of demographic groups.
    \item We study different data sampling approaches to mitigate the effects of extreme imbalance of training data in the context of embedding learning.
\end{itemize}

\section{Related work}
\subsection{Selfie-doc face recognition}
\begin{figure}
\centering
\includegraphics[width=.2\linewidth]{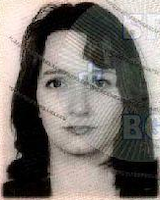}
\includegraphics[width=.2\linewidth]{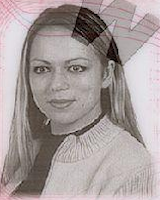}
\includegraphics[width=.2\linewidth]{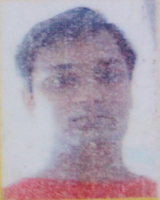}
\\
\includegraphics[width=.64\linewidth]{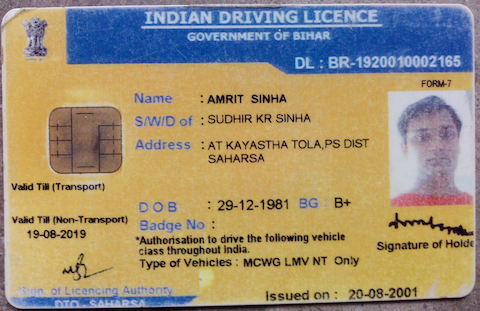}
\caption{Sample document photos (Bulgarian and Polish identity cards and an Indian driving licence). Security features and overall printing quality have a strong impact on the quality of the document photo. Source: https://commons.wikimedia.org/}
\label{fig:doc_samples}
\end{figure}

We consider a subproblem of face verification where a ``live'' or ``selfie'' photo is matched with a picture of an identity document. This problem poses unique challenges:
\begin{itemize}[nosep]
    \item Bi-sample data: only two samples are available per identity, making modelling of intra-class variations challenging.
    
    \item High number (many millions) of identities across a large range of demographic groups.
    
    \item Cross-domain data: While selfies are similar in character to images found in publicly available datasets such as {VGGFace2} or MegaFace, the document images differ in significant respects. A document photo is a photo of another photo that is embedded in the document with security features such as holograms or geometric patterns overlaid over the image. Security features vary across document types and some, such as holograms, also vary with the viewing angle and ambient light. Some examples are shown in Fig.~\ref{fig:doc_samples}. Furthermore, the printing quality and the physical condition of documents varies greatly and in some cases even a high quality photograph of the document will yield only a low quality image of the face as can be seen in the second row of Fig.~\ref{fig:doc_samples}.
\end{itemize}

Shi and Jain~\cite{docface2018} studied the related problem of matching selfie images with document images extracted from the document's chip. Extracting document facial images from the document's chip rather than from a digital photo of the document results in high quality document images with no occlusion by security features and no variation due to illumination and other imaging conditions. Furthermore, the variation of their dataset was limited by the fact that the authors only used Chinese Resident Identity Cards\footnote{https://en.wikipedia.org/wiki/Resident\_Identity\_Card}, so there was little ethnic variation in their data and the document images were fairly standardised.  However, even in this simplified setting, naively training the models on in-the-wild images proved unsuccessful. The authors treated the selfie-doc comparison problem as heterogeneous face recognition. They pretrained their model on a large-scale in-the-wild dataset using an adaptive margin softmax loss \cite{amsoftmax2018} before using it to initialise two sibling networks with the same architecture. Each of the sibling networks was then fine-tuned on the selfie-doc data using a novel max-margin pairwise score. 
Using an internal dataset of 10K selfies-doc pairs, the authors reported an improvement in the true acceptance rate (TAR) from 61.14\% to 92.77\% at 0.1\% false acceptance rate (FAR).
In a follow-up study~\cite{docfacep2019}, the authors modified their approach by sharing the high-level features in the sibling networks and used dynamic weight imprinting \cite{lowshot2018} for the transfer learning step. Shi and Jain reported more significant improvement using this approach compared to~\cite{docface2018}.


In another paper~\cite{bisample2018} on selfie-doc matching, the authors employed a 3-stage classification-verification-classification (CVC) strategy in which a model was first trained using a softmax classification approach on a small number of classes with multiple samples per class. The knowledge learned by this model was then transferred to the selfie-doc domain by a verification-like loss such as centre loss~\cite{centreloss2016} or triplet loss~\cite{facenet2015}. Finally, the network was further trained in a large-scale classification scenario which was made feasible due to a \textit{dominant prototype selection} strategy to mine a small number of dominant classes from the large set of available identities for each training batch. A private dataset, CASIA IvS, of more than 2 million identities was used for training. Although not explicitly mentioned, the document images in the CASIA IvS dataset also seem to be extracted from the chip of Chinese national identity cards. The authors reported an improvement in the TAR from 84.16\% at $10^{-5}$ FAR using the best baseline model to 91.92\% using their approach.


Recently, Albiero \etal~\cite{docselfie_adolescence2019} studied selfie-doc matching in the case where the document image is from early adolescence while the selfie is from late adolescence. An internal dataset with two versions of Chilean ID cards and over 260K identities was used. They compared multiple commercial off-the-shelf (COTS) and government off-the-shelf face recognition systems, a number of state-of-the-art CNNs, including ArcFace~\cite{arcface2018}, as well as the doc-selfie model of \cite{docfacep2019}. All baseline models performed sub-optimally when applied to the Chilean doc-selfie dataset; furthermore, the accuracy of the models drops with increasing age difference between the two images. Interestingly, the model that performed worst was the doc-selfie model~\cite{docfacep2019}, which was the only model that was fine-tuned on doc-selfie data. The authors hypothesize that this may due the differences between training and test data: the model was trained on mostly Asian identities and evaluated on Chilean identities. The authors showed how to improve the performance of ArcFace by fine-tuning on their dataset. To the best of our knowledge this is the only work in the doc-selfie domain that has examined demographic differentials. However, due to the limited ethnic variation in their dataset (all users were Chilean) they were unable to report on ethnic differentials. In contrast other work in the area, \cite{docselfie_adolescence2019} did not observe any significant gender differential in performance.

\subsection{Demographic performance differential}
A number of recent works have reported similar performance differentials across demographic groups in modern deep learning-based face recognition systems. Krishnapriya \etal~\cite{fr_acc_rel_race_2018} analysed three open source and one COTS algorithm and reported a large differential performance in terms of FAR between African-American vs. Caucasian groups. Cavazos \etal~\cite{where_are_we_otool2019} studied one pre-deep learning algorithm and three modern deep learning-based algorithms and  reported higher FAR in Asian vs. Caucasian identities. 

NIST face recognition vendor tests have found that face recognition algorithms exhibit varying accuracy across different demographic cohorts as early as 2002~\cite{frvt2002}. A recent NIST report~\cite{frvt2019_part3_demog} analysed many COTS and state-of-the-art algorithms and found widespread demographic differential across ethnicity, age, and gender. 

Despite widespread evidence of demographic bias, very few works have attempted to mitigate such bias in face recognition. For face detection, Amini \etal~\cite{latent_bias2019} used a debiasing variational auto-encoder to discover and mitigate hidden biases within training data. Wang \etal~\cite{racialfw2019} introduced a dedicated dataset, Racial Faces in the Wild, for measuring performance across four racial cohorts: African, Asian, Caucasian and Indian. Using this dataset, the authors measured the accuracy of four COTS and four state-of-the-art algorithms and reported that all algorithms showed different accuracy across the cohorts, with the non-Caucasian groups showing much lower accuracy. They showed that using a balanced training set can significantly improve the performance on non-Caucasian faces. Furthermore, \cite{racialfw2019} proposed an unsupervised domain adaptation framework using Caucasian faces as the source domain and other ethnic groups as the target domain and observed improved performance on the target groups, while assuming only access to unlabelled data for the target domain. 




\section{Selfie-doc dataset}
\label{sect:dataset}

We use a dataset of 6.8M image pairs, each consisting of a selfie taken using a smartphone camera or a computer webcam and a photo of a government issued ID document. This dataset was sampled from data collected at Onfido as part of its identity verification service. We call this the \emph{selfie-doc dataset} and it will be used for the experiments presented in Sect.~\ref{sect:experiments}. Due to data protection regulatory obligations we are not able to publicly release the dataset.


Document photos in this dataset are different from the photos stored on the chip of a biometric passport. The latter are high resolution images without any of the printing or security artifacts. However, they require NFC readers to be accessed and are not available for many document types.

We extract the document issuing countries and aggregate them into 30 groups, which are mapped to six continents.

\begin{description}[nosep]
\item[Europe]
France, Great Britain, Ireland, Italy, Latvia, Li\-thuania, Poland, Portugal, Roumania, Spain and Europe (rem)
\item[America]
Brazil, Canada, Colombia, USA, Venezuela and Americas (rem)
\item[Africa]
Nigeria, North Africa (Morocco, Tunisia, Algeria and Egypt), South Africa and Africa (rem)
\item[Asia]
China, India, Indonesia, Malaysia, Singapore, Thailand and Asia (rem)
\item[Oceania]
Oceania
\item[Unknown]
Unknown
\end{description}

The countries were chosen based on the available data. Countries with insufficient data to form a separate category where grouped together to form the remainder of a continent class. For simplicity, we call the resulting 30 groups again ``Countries''. Country and gender metadata were extracted from documents using a combination of automatic extraction and manual labelling. Automatic extraction used an in-house document classifier and document-specific OCR models to extract information from the document. The large Unknown class for both country and gender is due to unavailability of the data for various operational reasons\footnote{It would be possible to infer missing gender labels using a CNN. We did not follow this route in order to minimize processing of sensitive personal data and because gender is not the main focus of this paper.}.

As can be seen in Table \ref{tab:train_cont} our dataset exhibits extreme imbalance across continents with Africa making up only 0.5\% of the total number of samples and Asia making up less than 5\%. There is also a significant degree of gender imbalance among the samples with known gender with male users almost twice as much as female users.

All 6.8M image pairs were used for training. For evaluation we used a separate dataset of that were collected at a later time. The evaluation dataset contained 20K pairs with the same geographic distribution as the training set and 1K pairs from each country.

\begin{table}
\begin{center}
\begin{tabular}{lcccc}
\toprule
& Male & \!Female\! & \!\!Unknown\!\! & All \\
\midrule
Europe (EU) & \!29.0\%\! & 16.5\% & 15.5\% & 61.0\% \\
America (AM) & 9.2\% & 5.6\% & 0.3\% & 15.1\% \\
Africa (AF) & 0.3\% & 0.1\% & 0.1\% & 0.5\% \\
Asia (AS) & 2.4\% & 0.7\% & 1.6\% & 4.7\% \\
Oceania (OC) & 0.1\% & 0.1\% & 0.2\% & 0.3\% \\
Unknown (UN) & 0.0\% & 0.0\% & 18.3\% & 18.3\% \\
All & \!41.0\%\! & 23.0\% & 36.1\% & 100.0\% \\
\bottomrule
\end{tabular}
\end{center}
\caption{Composition of training data by continent and gender.}
\label{tab:train_cont}
\end{table}

\section{Method}
\label{sect:method}

To train the face recognition model we use triplet loss together with online semi-hard triplet mining and various domain-adapted data sampling strategies.

\subsection{Loss function and training}
We denote an image by $x$ and the corresponding feature embedding by $z = f(x)$; the embedding is computed using a neural network $f(\cdot)$ and we assume that the embeddings are normalized, \ie, $\|z\|_2 = 1$.

We train the network using triplet loss \cite{facenet2015},
\[
\mathcal L = \max\left( D_{ap}^2 - D_{an}^2 + \alpha, 0 \right) \,,
\]
where $(x_a, x_p, x_n)$ are triplets of images with the \emph{anchor} $x_a$ and the \emph{positive} image $x_p$ belonging to the same identity and the \emph{negative} image $x_n$ belongs to a different identity. The terms $D_{ap}^2 = \|f(x_a) - f(x_p)\|^2$ and $D_{an}^2 = \|f(x_a) - f(x_n)\|^2$ are squared Euclidean distances between the anchor and the positive/negative images, respectively, and $\alpha$ is a parameter enforcing a separation margin. In our experiments we set $\alpha=0.6$.

Recently, large-margin classification losses such as ArcFace~\cite{arcface2018} and SphereFace~\cite{sphereface2017} were shown to lead to better results on publicly available datasets than triplet loss, especially when combined with large-scale classification and weight-imprinting~\cite{lowshot2018}. However, publicly available datasets differ in important respects from the selfie-doc dataset we are using: (1) public datasets usually contain a large number of images per identity, while we have only two images per identity and (2) the identities in a public data are mutually exclusive, while the selfie-doc dataset contains repeated identities (approx.\ $2\%$ of identities have more than one image pair in the dataset).

Having only two images per identity implies that the anchor and positive images in a triplet already contain all the information about the identity that we are privy to, and techniques such as centre loss~\cite{centreloss2016} wouldn't provide additional benefits. In \cite{bisample2018} the authors worked with a similar dataset to ours, containing two images per identity, and they trained the network in stages: first using triplet loss and then using large-scale classification. In our case, because the identities are not mutually exclusive, we would need to perform significant data cleaning before large-scale classification can be applied. For triplet loss, however, the chance of constructing a triplet with the same identity for both positive and negative images is exceedingly low and thus the label noise has less impact on the training process.

The advantage of large-scale classification is that it provides a global view of the embedding space, while the triplet loss only sees three images at a time. We compensate for it by using a large batch size of $N=10240$ during the triplet selection process. We adopt a semi-hard triplet selection strategy: for each positive pair $x_a, x_p$ we consider all candidate $x_c$ images in the batch that violate the margin, i.e.,
\[
\|f(x_a) - f(x_p)\|^2 + \alpha > \|f(x_a) - f(x_c)\|^2\,,
\]
and we choose the negative image $x_n$ randomly among the candidates.

In the selfie-doc dataset the images in a pair $(x^s, x^d)$ come from different domains and we respect this during the triplet selection by choosing the negative image in the triplet from the same domain as the positive image, i.e. each pair $(x^s, x^d)$ is used to generate two triplets,
\[
(x^s, x^d, x^d_n)\,\text{and}\,
(x^d, x^s, x^s_n)\,.
\]
In one triplet the anchor is a selfie and the positive and negative images are document photos and in the other triplet the roles are reversed. This reflects the fact that we intend to use the model only for cross-modality matching.

Due to limited GPU memory we cannot perform optimization steps with batches of size $N$, and so after after all $2N$ triplets have been selected, we use minibatches of size $N_{train} = 32$ for the optimization step. 


\subsection{Evaluation}

We evaluate the performance of models using \emph{false acceptance rate (FAR)} and \emph{false rejection rate (FRR)}. Given a set $A = \{(x^s_i, x^d_i) \,:\, i=1,\dots, M\}$ of matching image pairs and a threshold $\theta$, FAR and FRR are defined as
\begin{align*}
\mathrm{FAR}(\theta) &= \frac{1}{M(M-1)} \sum_{i \neq j} \mathbb{1}\left(
\| f(x^s_i) - f(x^d_j) \|^2 < \theta \right) \,, \\
\mathrm{FRR}(\theta) &= \frac{1}{M} \sum_{i=1}^M \mathbb{1}\left(
\| f(x^s_i) - f(x^d_i) \|^2 \geq \theta \right)\,.
\end{align*}
To measure FAR we can also use two separate sets of selfie and document images with disjoint identities.

\subsection{Sampling strategies}

Given the extreme imbalance of the dataset as shown in Table~\ref{tab:train_cont}, it is expected that without specific sampling strategies to balance the training batches, a model trained on this dataset will not perform equally well across different demographic groups. In this paper we will study the effects of several sampling strategies to limit the performance differential caused by this data imbalance.

\emph{Sampling with fixed weights.} The simplest strategy is to sample each group with a specified probability. We specify the probability by assigning a weight to each group; the sampling probability is then given by the normalized weight vector. We consider an equal sampling strategy, where each group has weight 1, as well as adjusted weight selection with weights that are higher for some groups than others.

In both cases we sample a batch of size $N$ with the given probabilities and then select triplets from this batch. In particular, this means that each batch will contain samples from all groups and the positive and negative samples in a triplet can come from different groups. Note, however, that the semi-hard triplet mining process will eliminate easy cross-group triplets.

\emph{Sampling with dynamically adjusted weights.} The motivation for choosing higher weights for some groups is to emphasise groups with lower performance. However, we expect the relative ranking of groups to change during the course of the training: the worst-performing group is assigned the highest weight at first, which leads to improvements for this group and after a certain number of training steps another group might become the worst performing group. 

To account for this process we implemented a dynamic weight adjustment strategy. After a certain number of training steps we evaluate the performance of the model on a validation set and choose weights for a group inversely proportional to the performance of the model on the group.

In particular, we choose FAR as our performance measure. We set an acceptance threshold $\theta$ such that the overall FAR is at $10^{-5}$ and consider $\mathrm{FAR}_i(\theta)$, the within-group FAR for group $i$ at threshold $\theta$, as the performance measure. The weights are set as 
$
\tilde{w}_i = \mathrm{FAR}_i(\theta)^{\lambda}\,,
$
and we choose $\lambda = \log_{10} 4$. This implies that a 10-fold increase in FAR yields a 4-fold increase in the weights.

To smooth the training process we use exponential averaging, i.e., the weights for epoch $t$ are computed as
\[
w_i(t) = \alpha \tilde w_i(t) + (1-\alpha) w_i(t-1)\,,
\]
where $\alpha=0.2$ in our experiments.

\emph{Homogeneous batches.} We also consider using batches for triplet selection that contain samples from only one group. In this strategy we choose the group using predefined or dynamically adjusted weights and then select all $N$ samples from that group. The idea is to make batches more homogeneous and thus increase the potential number of hard samples for triplet selection.

\section{Experiments}
\label{sect:experiments}

\subsection{Implementation details}

For data preprocessing we use MTCNN \cite{mtcnn2016} to detect the face, which is then resized to $112 \times 112$ pixels. Following \cite{arcface2018} we utilize for the embedding network a ResNet-100 architecture with a BN \cite{batchnorm2015}-Dropout \cite{dropout2014}-BN-FC structure after the last convolutional layer.

The network is first trained on MS1MV2 \cite{arcface2018}, a semi-automatically refined version of the MS-Celeb-1M dataset \cite{ms1m2016}. The ethnic diversity of MS1MV2 is as follows~\cite{racialfw2019}: Caucasian 76.3\%, Asian 6.6\%, Indian 2.6\%, and African 14.5\%. We use the weights from the InsightFace repository as the starting point for our experiments\footnote{https://github.com/deepinsight/insightface}.

\subsection{Fine-tuned model}

\begin{figure}
\centering
\includegraphics[width=\linewidth]{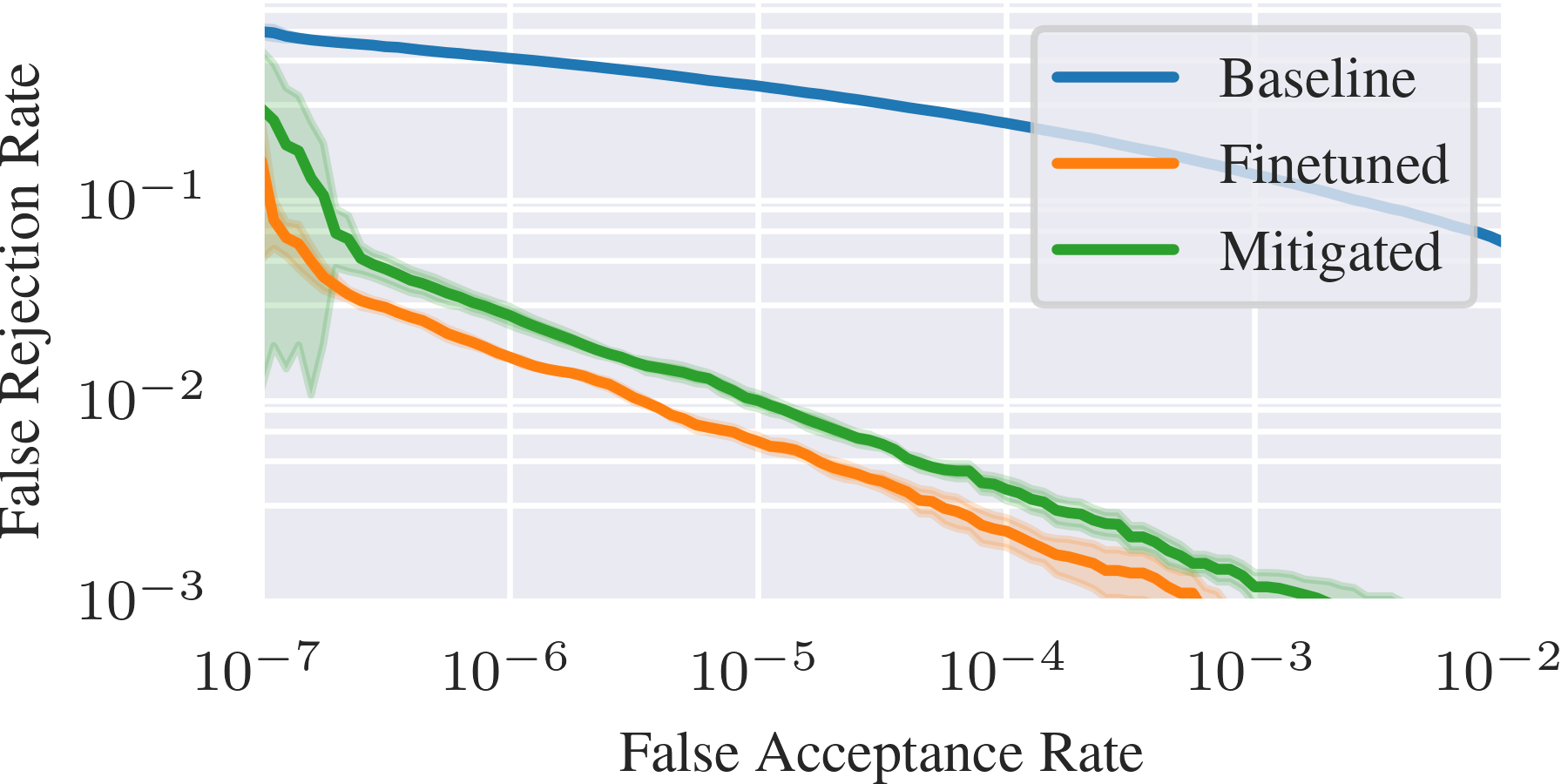}
\caption{ROC curves for the baseline models. The model trained on MS1MV2 exhibits poor performance on the selfie-doc dataset, while the finetuned model has less than 1\% FRR at $10^{-5}$ FAR. The mitigated model is trained with country-based adjusted sampling and has slightly higher FRR. We averaged the ROC curve over five different val/test splits and show the standard deviation. We see in the results are very stable in the region $10^{-4}$ to $10^{-6}$ FAR.}
\label{fig:baseline_roc}
\end{figure}

The baseline ResNet-100 model was trained on the publicly available MS1MV2 dataset and achieves 99.77\% on the LFW benchmark~\cite{lfw2008} and 98.47\% on the MegaFace~\cite{megaface2016} benchmark, but it performs poorly when evaluated on selfie-doc pairs. As can be seen in Fig.~\ref{fig:baseline_roc}, at $10^{-5}$ FAR its FRR is above 20\%. 
Because document images are visually significantly different from the images found in MS-Celeb-1M, this drop is expected and in line with  \cite{bisample2018} and \cite{docfacep2019}.

The finetuned model was trained using triplet loss with triplets selected across 10K candidates and an optimization batch size of 32. We used the Adam optimizer with an initial learning rate $10^{-5}$ which was gradually decreased to $10^{-7}$. The model was trained for 2.7M steps.

After finetuning the ResNet-100 model on selfie-doc data using triplet loss the performance improves significantly to $0.6\%$ FRR at $10^{-5}$ FAR. We will use this finetuned ResNet-100 model as the starting point for exploring various sampling strategies.

\subsection{Differential performance by continent}
\label{sect:continent}

\begin{figure}
\centering
\includegraphics[width=\linewidth]{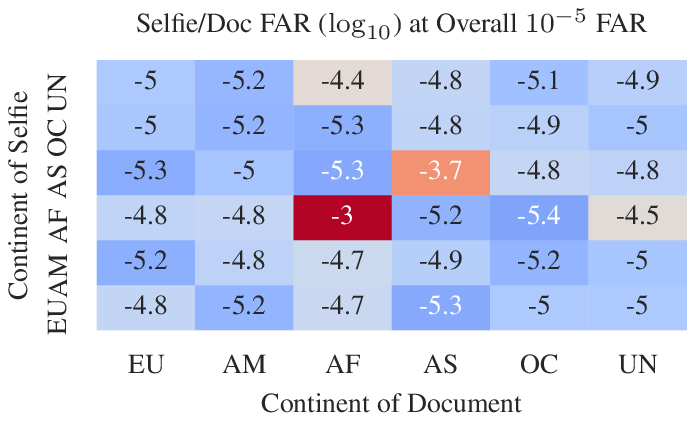}
\caption{FAR for the finetuned model by continent. Each cell is calculated using 1M selfie-doc pairs generated using 1K selfies and 1K documents. The diagonal contains comparisons of selfies and documents from the same continent and therefore higher FARs than off-diagonal entries.}
\label{fig:baseline_continent_far}
\end{figure}

While the overall performance is satisfactory, the picture becomes less bright when we interrogate the performance by the continent in which the document was issued. In Fig.~\ref{fig:baseline_continent_far} we see that the FAR for African selfie-doc pairs is 100 times (2 orders of magnitude) higher than the overall FAR, while the FAR for Europe and America is only 1.5 times higher (0.2 orders of magnitude) than the overall FAR. This is not surprising given the imbalance in the training set as shown in Table~\ref{tab:train_cont}. Only 0.5\% of all training samples are from African documents and only 4.7\% of training samples are from Asian documents. 

\begin{figure}
\centering
\includegraphics[width=\linewidth]{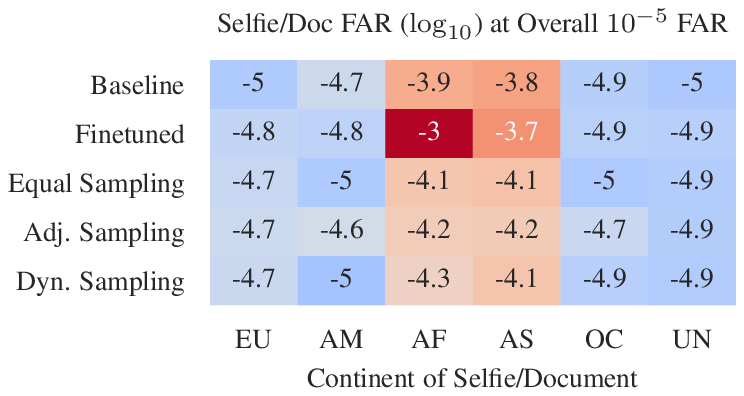}
\caption{FAR for selfie-doc pairs from the same continent for various continent-based sampling strategies. All three strategies reduce the performance differential with small differences between them.}
\label{fig:continent_strategies_far}
\end{figure}

We can make another interesting observation by comparing the continent-specific FARs for the baseline and the finetuned model in Fig.~\ref{fig:continent_strategies_far}. While the finetuned model has a greatly reduced FRR at an overall FAR of $10^{-5}$, it does exhibit greater differences in FAR across groups than the baseline model. Thus, we can conclude that finetuning the model on the selfie-doc dataset improved the FRR, but also increased the demographic differential for FAR.

We try various strategies to address the FAR imbalance. 
\begin{itemize}[nosep]
\item Equal sampling---we sample each continent equally often during training
\item Adjusted sampling---we use weighted sampling and assign EU, AM, OC and UN weight 1 and AF, AS weight 3.
\item Dynamic sampling---we dynamically vary sampling weights during training based on continent-specific FARs as measured on a validation set; see Sect.~\ref{sect:method}.
\end{itemize}
The impact of these strategies can be seen in Fig.~\ref{fig:continent_strategies_far}. The adjusted sampling strategy performs slightly better than sampling with equal weights. We speculate that the reason for this is that samples from Europe, America and Oceania are 
visually more similar
and so equal sampling across all continents actually undersamples both Africa and Asia. On the other hand it can be legitimately argued that the choice of weights (3 and 1) in the adjusted sampling strategy is ad hoc and so the dynamic sampling strategy is a more principled alternative and provides a similar level of bias mitigation.

\begin{figure}
\centering
\includegraphics[width=\linewidth]{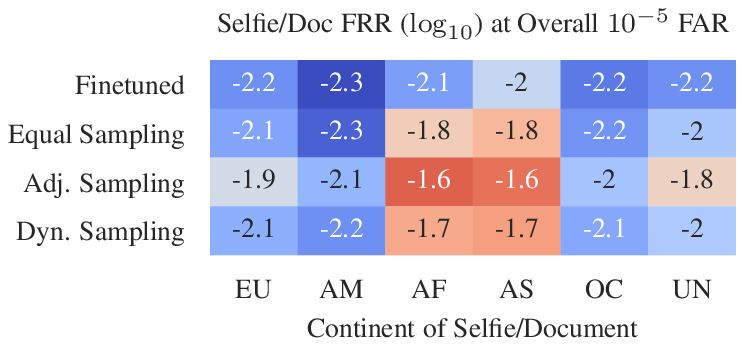}
\caption{FRR for various continent-based sampling strategies. We omit the baseline, because its FRR is above $10^{-1}$.}
\label{fig:continent_strategies_frr}
\end{figure}

\emph{Homogeneous batches.} When training with homogeneous batches of size $N=2048$, chosen with equal weight for all continents, we observe that the within-continent FAR swings dramatically during the training and there is an exchange taking place between the FARs for Africa and Asia. We hypothesize that because of the homogeneity of triplets the model has no opportunity to learn how to simultaneously embed samples from all continents and thus, based on the samples it has seen most recently it will have good performance on one group at the expense of another group. This can be seen as another manifestation of catastrophic forgetting~\cite{forgetting2019}.

\begin{figure}
\centering
\includegraphics[width=\linewidth]{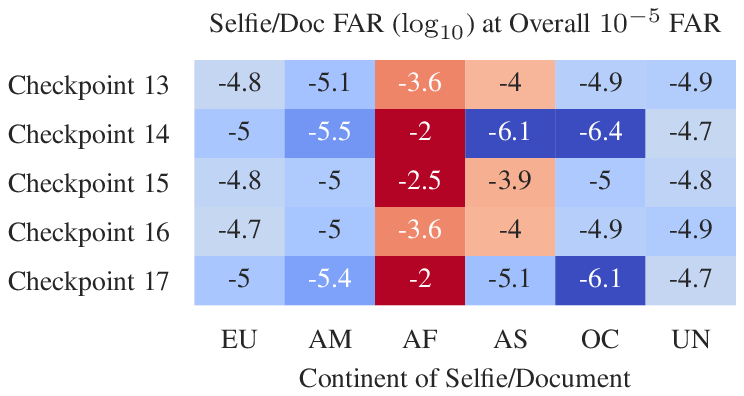}
\caption{FAR for selfie-doc pairs during the training with homogeneous batches, each batch contains samples from only one continent.}
\label{fig:homogeneous_sampling_checkpoints_far}
\end{figure}

\subsection{Differential performance by country}

\begin{figure}
\centering
\includegraphics[width=\linewidth]{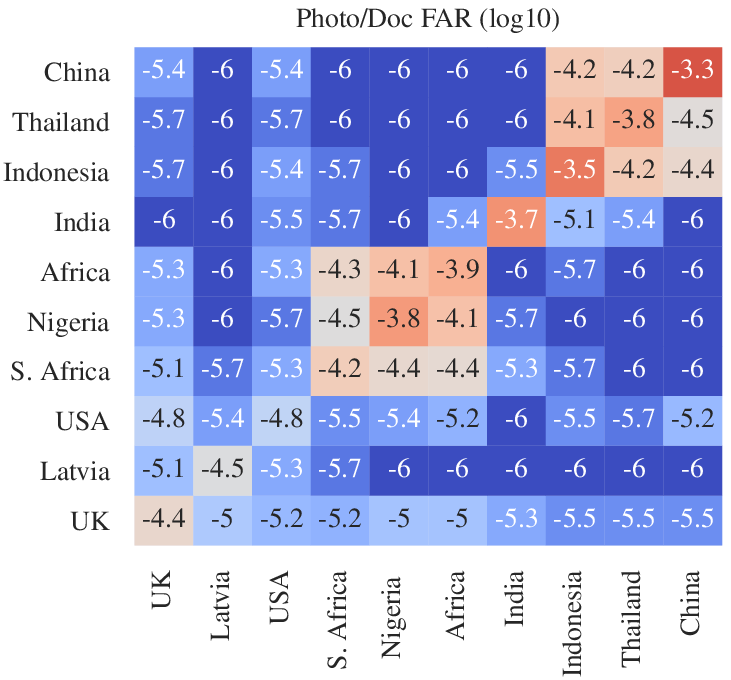}
\caption{FARs for the continent-based dynamic sampling model by country for selected countries. We see that the $10^{-4.1}$ FAR for Asia becomes $10^{-3.3}$ within China while being as low as $10^{-4.5}$ for Thailand/China pairs.}
\label{fig:best_cont_by_country}
\end{figure}

We can take an even more fine-grained look at the performance of our model and consider the FARs within single countries (as defined in Sect.~\ref{sect:dataset}). In Fig.~\ref{fig:best_cont_by_country} we see the by country FAR for the model trained with dynamic sampling by continents. Two features are visible: there are clusters of countries in Africa and Asia with high FARs off the diagonal, showing that there are significant similarities of images across countries. On the other hand, Asia seems to split into (at least) two separate clusters with India forming one cluster and Indonesia, Thailand and China the second cluster.

The matrix in Fig.~\ref{fig:best_cont_by_country} is almost symmetric, meaning the FAR for matching selfies from country A with documents from country B is almost the same as the other way around. The degree of symmetry suggests that FAR differences in logarithmic space of $0.3$ and larger, corresponding to a two-fold increase in FAR, are genuine and not due to noise or sample selection.

We compare three different sampling strategies for mitigating the country-based performance differential.
\begin{itemize}[nosep]
\item
Continent-based dynamic sampling (Sect.~\ref{sect:continent})
\item
Adjusted sampling---we use weighted sampling and assign countries from Africa, Asia and America except for USA and Canada weight 4 and all remaining countries weight 1.
\item
Dynamic sampling---we dynamically vary sampling weights based on country-specific FARs; see Sect.~\ref{sect:method}.
\end{itemize}

\begin{figure*}
\centering
\includegraphics[width=\linewidth]{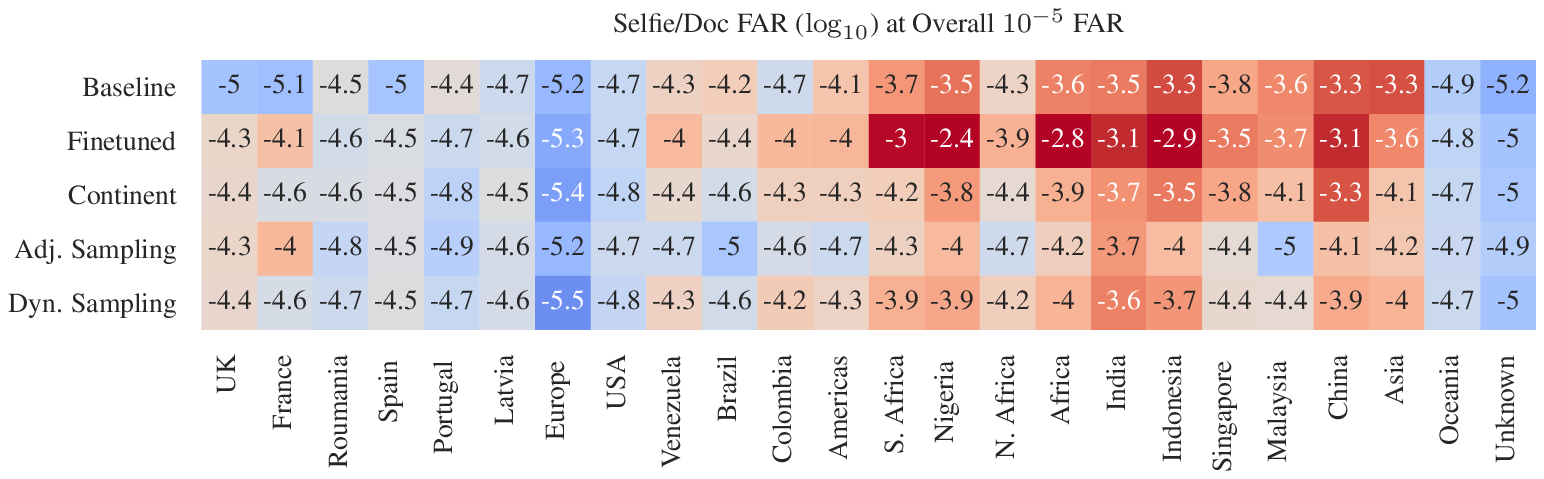}
\caption{FAR for selfie-doc pairs from the same country for various sampling strategies. `Continent' is dynamic continent-based sampling; `Adj. Sampling' is country-based sampling with fixed weights and `Dyn. Sampling' is country-based sampling with dynamically adjusted weights.}
\label{fig:country_far}
\end{figure*}

The results we obtain are mixed. First, as can be seen in Fig.~\ref{fig:country_far}, country-specific sampling outperforms continent-specific sampling for most countries. We attribute this to the fact that countries are unequally distributed within continents and that equalising the sampling frequency of each country is beneficial for the overall training.

On the other hand, the dynamic sampling strategy performs worse than a sampling strategy with fixed weights. The reasons for this are not entirely clear, but might be related to the fact that we don't have enough samples for each country and hence additional sampling from a given country leads to overfitting rather than improved performance.


Figure \ref{fig:baseline_roc} also compares ROC curve of the finetuned model with that of the country-based adjusted sampling strategy, as a typical example. We observe that the improvement in geographic bias comes at the cost of a slightly higher FRR at a given overal FAR. However, this increase is small compared to the scale of improvement in country-specific FARs (Figure \ref{fig:country_far}).

\subsection{Differential performance by gender}

We have concentrated on reducing performance differentials by continent and country, since that is where the baseline model showed the largest performance differential.

\begin{figure}
\centering
\includegraphics[width=\linewidth]{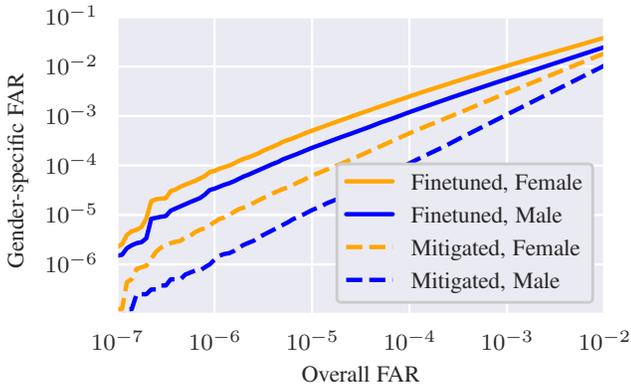}
\label{fig:gender_differential}
\caption{Gender-specific FARs at a given overall FAR. We compare the finetuned model, exhibiting large country-based FAR differentials, with the country-based adjusted sampling model. While the latter model has better absolute performance and better gender-specific performance, it also has greater gender differential.}
\end{figure}

However, when we compare the gender performance in Fig.~\ref{fig:gender_differential} of the finetuned model with the country-based adjusted sampling model, we can make the following observation. The finetuned model has a slight gender-bias: at a given overall FAR, the female FAR at that threshold is about twice as large as the male FAR. But for the country-based model, while the absolute performance for both genders is much better, i.e. the gender-specific FAR is smaller, the differential between the two genders is also larger. In other words, while both male and female FARs are lower after mitigation, the gap between them is wider after mitigation compared to before. This could be connected to gender-specific data imbalance in the training set, and it is interesting to see how mitigating a performance differential with respect to one grouping can lead to an increased performance differential in a complimentary grouping.

\subsection{Geometry of the embedding space}

We visualize changes to the embedding space during training with mitigation strategies by  comparing UMAP~\cite{umap2018} projections of embeddings for the finetuned model and the continent-based model with dynamically sampled weights (rows 2 and 5 in Fig.~\ref{fig:continent_strategies_far}). We can observe that continent-based sampling allowed the model to create more connections between African samples (green) and other continents. For the finetuned model the African samples form a tight cluster on their own, disconnected from the main clusters, but for the continent-based model the African cluster is more closely integrated with the other clusters and is less dense, implying higher separability within this group.

\begin{figure}
\centering
\includegraphics[width=.49\linewidth]{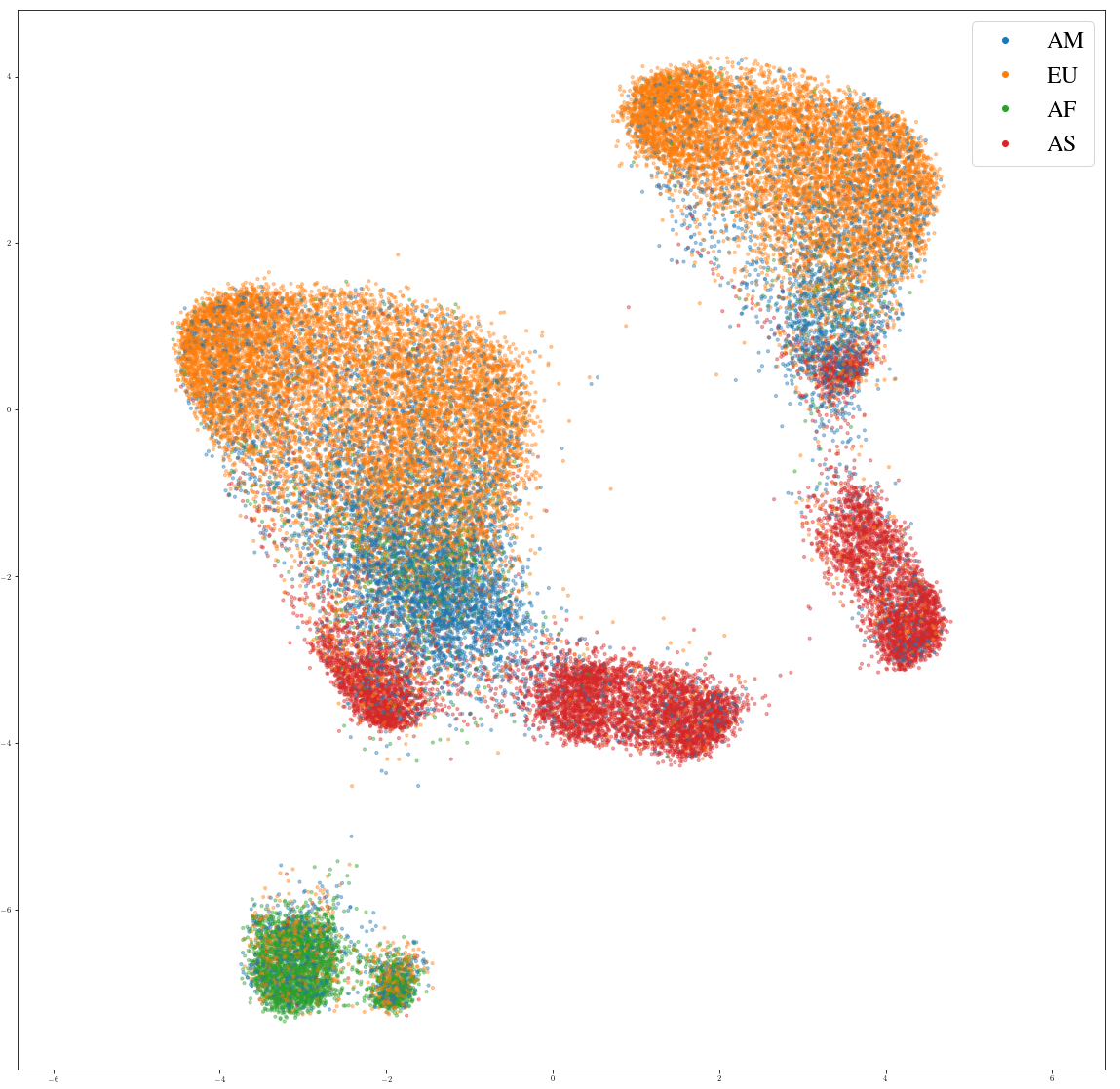}
\includegraphics[width=.49\linewidth]{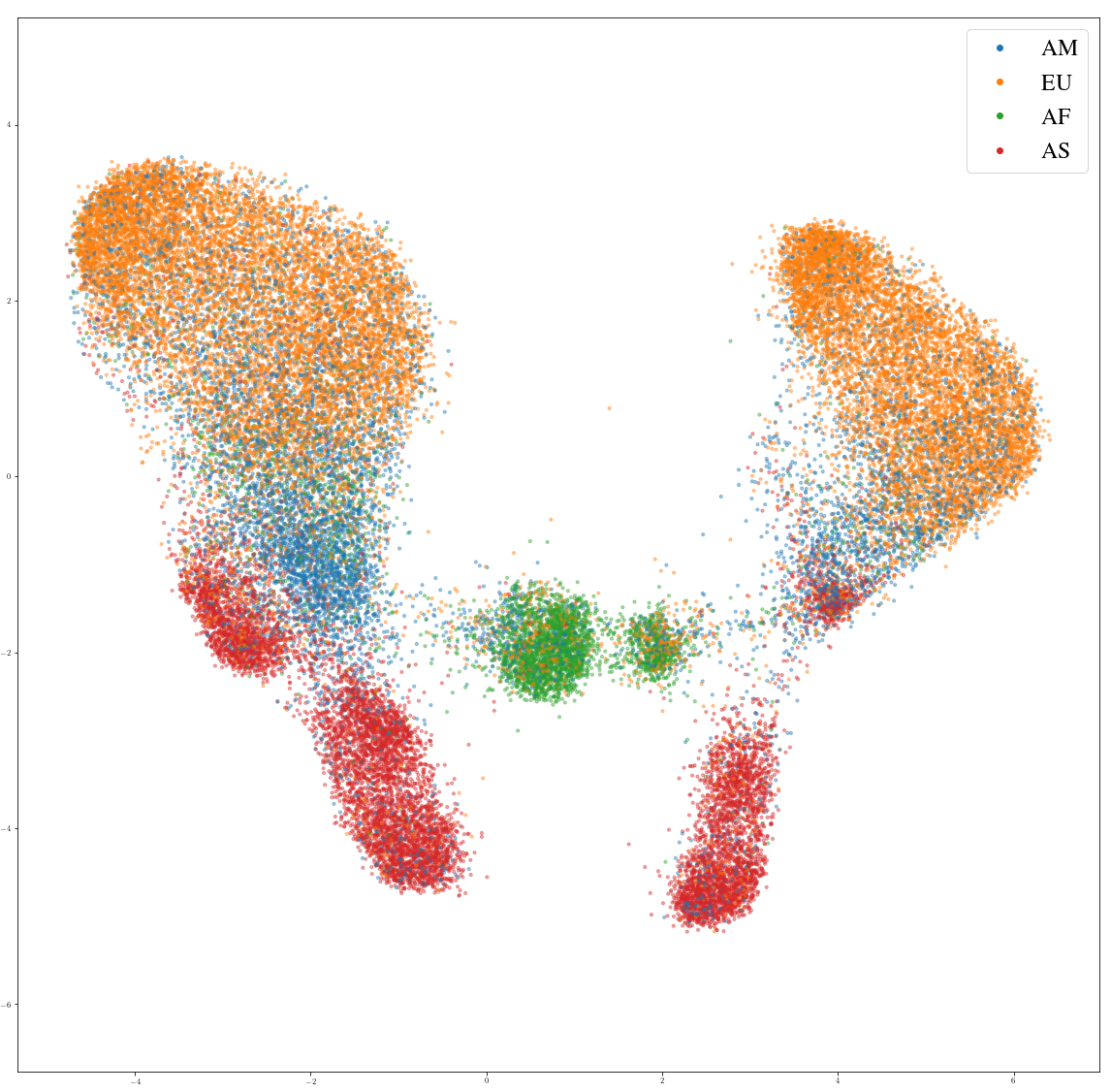}
\caption{UMAP projection of embeddings for the finetuned (left) and continent-based dynamically sampled models (right). The two large clusters correspond to gender and Asia is split into two subclusters corresponding to India and the rest of Asia.}
\label{fig:umap_before_mitigation}
\end{figure}



\section{Discussion}

There are some learnings we can take away from the experiments presented above.

A balanced dataset is not a requirement to reduce performance differentials in a face recognition system. Even though only 0.5\% of the images in selfie-doc dataset are from Africa, we were able to reduce the FAR differential between Africa and Europe from a factor of 63 to a factor of 2.5---a 25.2-fold reduction; see Fig.~\ref{fig:continent_strategies_far}. Weighted sampling strategies, whether with static or dynamic weights, can, to a certain extent, cope even with extremely imbalanced datasets.

On the other hand, having fine-grained labels for the training set is an advantage. In the absence of country- or continent-level labels one can estimate the labels using a trained classifier---it is feasible to train a country classifier on document images, for selfies one would have to resort to intrinsic facial characteristics such as apparent ethnicity. An alternative approach is to use unsupervised clustering methods to find clusters that are natural to the data and the model performance. Such a strategy was applied in \cite{latent_bias2019} for training a face classifier.

For sampling strategies with dynamic weight selection to be successful it is important to have a validation set with clean labels. Neural network training is usually robust to a certain level of label noise \cite{noise2017}, but a noisy validation set will impact the sampling strategy. 

We do not claim that the observed performance differentials are caused solely by ethnicity. Other factors, such as differing camera quality and different types of security features, can impact performance as well. Disentangling cause from correlation remains an open question.

Removing demographic performance differentials in face recognition is a multi-objective optimization problem. Focusing on one objective, such as country-specific FAR, can have adverse effects on other objectives, such as
gender differentials, although we observed that gender-specific FARs improved.
Similarly there might be a trade-off between reducing FARs and FRRs. How to properly balance competing demographic groups and competing performance measures remains the subject of future research.

{\small
\bibliographystyle{ieee}
\bibliography{bibliography}
}

\end{document}